\documentclass{article}

\usepackage{arxiv}

\usepackage[utf8]{inputenc} % allow utf-8 input
\usepackage[T1]{fontenc}    % use 8-bit T1 fonts
\usepackage{hyperref}       % hyperlinks
\usepackage{url}            % simple URL typesetting
\usepackage{booktabs}       % professional-quality tables
\usepackage{nicefrac}       % compact symbols for 1/2, etc.
\usepackage{microtype}      % microtypography
\usepackage{lipsum}		% Can be removed after putting your text content
\usepackage{natbib}
\usepackage{doi}
\setcitestyle{numbers,square}
\usepackage{amsmath,amssymb,amsfonts}
\usepackage{graphicx,color}
\usepackage{textcomp}
\usepackage{xcolor}
\usepackage{algorithm,algorithmic}
\usepackage{threeparttable} % to use table notes
\usepackage{multirow}

\title{ExpressNet-MoE: A Hybrid Deep Neural Network for Emotion Recognition\thanks{{This paper is currently under review at a publisher.}}}

\author{ \href{https://orcid.org/0000-0002-6967-3807}{\includegraphics[scale=0.06]{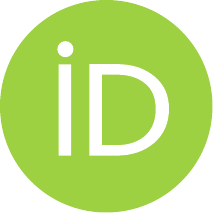}\hspace{1mm}Deeptimaan Banerjee} \\
	Computer Science \& Engineering,\\
	University of Colorado Denver,\\
	Colorado, CO 80204, \\
	\texttt{deeptimaan.banerjee@ucdenver.edu} \\
	\And
	Prateek Gothwal \\
	Computer Science \& Engineering,\\
	University of Colorado Denver,\\
	Colorado, CO 80204, \\
	\texttt{prateek.gothwal@ucdenver.edu}
	\AND
	\href{https://orcid.org/0000-0002-7446-4639}{\includegraphics[scale=0.06]{orcid.pdf}\hspace{1mm}Ashis Kumer Biswas} \\
	Computer Science \& Engineering,\\
	University of Colorado Denver,\\
	Colorado, CO 80204, \\
	\texttt{ashis.biswas@ucdenver.edu}
}

\date{October 10, 2025}

\renewcommand{\shorttitle}{ExpressNet-MoE: A Hybrid Deep Neural Network for Emotion Recognition}

\hypersetup{
pdftitle={ExpressNet-MoE: A hybrid deep neural network for emotion recognition},
pdfsubject={AI},
pdfauthor={Deeptimaan Banerjee, Prateek Gothwal, Ashis Kumer Biswas},
pdfkeywords={Facial Emotion Recognition, 
Mixture of Experts},
}

\begin{document}
\maketitle

\begin{abstract}
In many domains, including online education, healthcare, security, and human-computer interaction, facial emotion recognition (FER) is essential. Real-world FER is still difficult despite its significance because of some factors such as variable head positions, occlusions,  illumination shifts, and demographic diversity. Engagement detection, which is essential for applications like virtual learning and customer services, is frequently challenging due to FER limitations by many current models
%, due to their performance on real-world data
. In this article, we propose ExpressNet-MoE, a novel hybrid deep learning model that blends both Convolution Neural Networks (CNNs) and Mixture of Experts (MoE) framework, to overcome the difficulties. Our model dynamically chooses the most pertinent expert networks, thus it
%the MoE module 
aids in the generalization and providing flexibility 
%of the 
to model 
across a wide variety of datasets. Our model improves on the accuracy of emotion recognition by utilizing multi-scale feature extraction to collect both global and local facial features. ExpressNet-MoE includes numerous CNN-based feature extractors, a MoE module for adaptive feature selection, and finally a residual network backbone for deep feature learning. To demonstrate efficacy of our proposed model we evaluated on several datasets, and compared with current state-of-the-art methods. Our model achieves accuracies of 74.77\% on $\text{AffectNet}_7$, 72.55\% on $\text{AffectNet}_8$, 84.29\% on RAF-DB, and 64.66\% on FER-2013. 
%Thus, improving on the performance of many models demonstrating state-of-the-art performance using the AffectNet dataset for testing. 
The results show how adaptive our model is and how it may be used to develop end-to-end emotion recognition systems in practical settings. Reproducible codes and results are made publicly accessible at \url{https://github.com/DeeptimaanB/ExpressNet-MoE}.
\end{abstract}

% keywords can be removed
\keywords{Adaptive Learning, 
Convolution Neural Networks, 
%Deep Learning, 
%Emotion Classification, 
Facial Emotion Recognition, 
Mixture of Experts, 
Generalization}

\section{Introduction}
\label{sec:intro}
Facial Emotion Recognition (FER) has become a major component of virtual education \cite{farman2023facial}, security \cite{ghadekar2024emosecure}, medical industry \cite{bohi2024novel}, \cite{huang2023emotion}, and human-computer interfaces. The ability to sense and respond to human emotions has created new opportunities for developing more flexible and intelligent systems. However, since human facial expressions are inherently complex, FER in real-world scenarios is a difficult task. Variations in head position, occlusion, lighting conditions \cite{farman2023facial}, \cite{sajjad2023comprehensive}, \cite{zhang2023dual} and demographic variability frequently cause significant performance issues in emotion detection frameworks. 
%Engagement detection, including recognizing whether or not a user actively engages or disengages, is particularly important for functions such as virtual education, where the involvement of students can significantly influence the learning outcomes, and for customer support, where detection of customer engagement can enhance the quality of support. Engagement detection greatly draws by the current emotion of a user and classifying emotions with accuracy will further aid these systems.

%Due to current research regarding Deep Learning networks, 
The performance of FER frameworks has improved significantly over time. There have been multiple ways for FER and its classification, including Local Binary Patterns (LBP) and Histogram of Oriented Gradients (HOG), combined with traditional machine learning algorithms such as Support Vector Machines (SVM) \cite{shan2009facial}, \cite{shan2005robust} and Random Forests \cite{nazir2018facial}, \cite{salhi2012fast}. Convolution Neural Networks (CNNs) have played a pivotal role in learning hierarchical feature representations \cite{khodaverdian2021shallow}.
%(Ashis revised this part to add reviewer1:paper3) from facial images and capture both high-level semantics and low-level spatial details. 
Moreover, transfer learning-based models have further enhanced recognition of complex emotional expressions due to the fact that they are pre-trained on multiple datasets. However, there are still many challenges in the machine learning development pipeline, such as unbalanced datasets and intra-class variations, leading to limited generalization.

Hence, we propose ExpressNet-MoE, a novel Deep CNN architecture. It is a hybrid deep learning model that combines CNNs with a layer of mixture of experts (MoE) to overcome the aforementioned problems. This method dynamically selects the relevant expert networks for each input, while improving its versatility and generalization over a number of datasets. ExpressNet-MoE captures simultaneously both global and fine-grained facial features and improves the accuracy of the emotion recognition task. The hybrid architecture of the model offers the most adaptive feature selection capability, and the choice of multiple CNN-based feature extractors makes it a powerful yet flexible approach.

The contributions of our research include the following.
\begin{itemize}
    \item \textbf{Adaptive feature learning}: Many existing solutions employ static models offering only a fixed set of features. To overcome the limitations of static feature extraction models, ExpressNet-MoE chooses a Mixture of Experts (MoE) system that emphasizes the best expert network in the model for each individual input.
    \item \textbf{Multi-scale feature extraction}: Our model extracts both global and fine-grained facial expression characteristics using CNNs with different filter sizes. The model's capacity to identify minute emotional variations across datasets is enhanced by the choice of this hybrid method.
    \item \textbf{Improved generalization across multiple datasets and real-world scenarios}: Our testing results showed that the model improves the generalization capability across different datasets due to its hybrid architecture. This allows the model to handle real-world issues such as illumination, occlusion, and demographic diversity. 
    %Therefore, it adds the improved This improves on the current knowledge of existing FER systems.
\end{itemize}

We have evaluated ExpressNet-MoE on three benchmark FER datasets which are AffectNet \cite{mollahosseini2017affectnet}, Real-world Affective Faces Database or RAF-DB  \cite{li2017reliable,li2019reliable}, and FER-2013 \cite{goodfellow2013challenges}. Each of these datasets has their own set of unique characteristics and problems. %AffectNet is perfect for training and validation since it provides a vast and varied collection of facial emotions from real-world situations. RAF-DB is reputed for its precise labeling; therefore, it is a good choice for testing the model’s generalizability as well. FER-2013 is a good dataset to test the model’s performance under constrained circumstances as it contains greyscale images and poor resolution and possible mislabeling problems. We evaluate our model on various datasets to show how it responds to variations in face expressions, environmental factors, and demographics.

Furthermore, ExpressNet-MoE employs deep feature representations, which offers more data-driven manner due to its dynamic MoE and transfer learning architecture to comprehend user emotions than conventional fixed CNN-based solutions. %This is especially helpful in applications like virtual classrooms, where real-time emotion recognition can improve education quality, and student engagement, where understanding a student's level of involvement can also personalize their learning experience.

The remainder of this article is organized as follows: Section \ref{sec:related_works} reviews related works related to current state-of-the-art emotion recognition methods. Section \ref{sec:dataset} provides the datasets utilized for both training and evaluation of our proposed model, along with comparing its performance with existing work. In Sections \ref{sec:proposed_method} and \ref{sec:model_architecture}, we present our proposed methodology and ExpressNet-MoE architecture, and its associated machine learning pipeline needed for end-to-end learning. Section \ref{sec:results} illustrates the experimental results of the proposed method separately for each of the datasets considered in this study. A comparative analysis is described in Section \ref{sec:comparison}. Section \ref{sec:discussion} provides an insight on the model performance and efficacy considerations. Finally, section \ref{sec:conclusions} summarizes the study, offers our conclusion and elaborates on future work.

\section{LITERATURE REVIEW}
\label{sec:related_works}
Understanding facial emotions has become a key element in emotion detection frameworks, particularly in applications such as human-computer interaction and online learning. Recent research focuses have been on the improvement of FER model accuracy, scalability, and generalization. Zhang et al. \cite{zhang2023dual} proposed a Dual-Direction Attention Mixed Feature Network (DDAN-MFN), which integrates a Mixed Feature Network (MFN) with a Dual-Direction Attention Network (DDAN). The MFN uses convolution and bottleneck layers along with MixConv, which uses several kernel sizes. The DDAN has an independent dual-direction attention head to capture long-range dependence, which significantly enhances the ability of the model to highlight the enlightening features of the FER undertaking.

Bhati et al. \cite{bhati2025generalized} proposed the Generalized Zero-Shot Convolution Neural Network (GZS-ConvNet), which aims at addressing the problem of generalization in FER systems. This architecture is designed to detect unseen facial expression using a sophisticated adaptation mechanism that exhibits high performance on a variety of datasets including FER-2013, AffectNet, and RAF-DB. GZS-ConvNet's ability to perform zero-shot categorization makes it a valuable tool for dynamic real-world applications where new expressions are likely to appear. Similarly, Bohi et al. \cite{bohi2024novel} proposed ConvNeXt, a CNN architecture that surpasses the classical CNN model and Transformer-based systems. The architecture uses deep convolutions, depthwise convolutions, and a modified activation function. The model demonstrates competitive performance on the AffectNet dataset.
Face2Nodes, a graph-based FER system presented by Jiang et al. \cite{jiang2023face2nodes}, uses dynamic relation-aware graph convolutions to characterize spatial and relational relationships between face regions, allowing for more structured and expressive emotion representations. Their work demonstrates strong performance on the RAF-DB dataset.

Uniyal et al. \cite{uniyal2024analyzing} in their research explored techniques to avoid overfitting and improve generalization. They performed research on deep convolution layers for emotion categorization using strategies such as max pooling, dropout, and batch normalization. Their methods underline the importance of feature extraction and regularization to achieve high accuracy on large datasets identical to AffectNet and FER-2013. Similarly, Savchenko et al. \cite{savchenko2022classifying} focused on the use of ensemble models, CNNs like VGG and ResNet, for emotion and engagement in distance learning.

Multimodal data and attention mechanisms have been successfully integrated to improve FER systems. In order to enhance emotion recognition, Sun et al. \cite{sun2024multimodal} suggested a multi-modal sentimental privileged information embedding (IA-MTM) that integrates audio and picture characteristics. By using both visual and aural input, their model improves FER performance by using ResNet18 for image feature extraction and an audio-decoding network to create a shared feature space. To improve FER models, Wang et al. \cite{wang2024qcs} presented the Cross Similarity Attention (CSA) mechanism. In order to solve the problem of class imbalance and enhance model performance on fine-grained information, the CSA pushes distinct emotion classes apart and brings similar ones closer together.

A CNN-based architecture for FER was proposed by Rajavenkatanarayanan et al. \cite{rajavenkatanarayanan2018monitoring} that effectively classifies facial expressions into positive, negative, and neutral classes using global average pooling and residual depth-wise separable convolutions. In order to improve feature extraction and learning stability, Roy et al. \cite{roy2024resemotenet} developed ResEmoteNet, which combines residual networks, convolution blocks, and Squeeze and Excitation (SE) blocks. Their model's capability to mitigate the impacts of vanishing and exploding gradients is demonstrated by its strong performance on a number of datasets.

Huang et al. \cite{huang2023study} developed the SE-ResNet architecture, which combines SENet with ResNet. The model successfully captures the importance of each channel by integrating the SE block into the ResNet architecture, guaranteeing improved learning during training and increased accuracy in FER tasks. Similarly, Dewan et al. \cite{dewan2019engagement} examined how machine learning models such as Bayesian classifiers and C4.5 trees are used to understand affective states and involvement in online learning environments. These algorithms analyze motions and facial expressions to help improve the tracking of student attention.

In order to improve feature representation, Halim et al. \cite{halim2023facial} created a Deep Convolution Neural Network with Convolution Block Attention Module (DCNN-CBAM), which makes use of attention processes. This design, which uses deep layers and convolution blocks and is tailored for the $\text{AffectNet}_8$ dataset, has shown promise in identifying students' emotional states during online instruction. 

%In conclusion, FER systems have significantly improved with the use of cutting-edge methods including residual learning, multimodal data, and attention processes. From engagement tracking in online learning to emotion detection in healthcare, these advancements guarantee that facial expression recognition algorithms can generalize effectively across a range of contexts. These research works show how FER architectures are constantly evolving, which emphasizes how crucial these models are becoming to human-computer interaction, especially in applications that demand precise, real-time emotion recognition.

Transformer-based architectures have also made their way into FER because of their capacity to recognize long-range correlations in face characteristics. For emotion classification, Roka and Rawat \cite{roka2023fine} investigated a Vision Transformer (ViT) model that had been pre-trained on ImageNet-21k and refined it using the AffectNet dataset. Their research highlights how crucial large-scale data and data augmentation are to addressing FER's class imbalance. Similarly, Huang et al. \cite{huang2021facial} introduced a hybrid framework FER-VT, which combines CNNs with attention mechanisms. It introduces Visual Transformer Attention (VTA) for high-level semantic representation and Grid-Wise Attention (GWA) for low-level feature extraction. The increasing trend toward transformer-based solutions in FER systems is shown in both studies. These methods overcome the drawbacks of traditional CNNs, especially their inability to accurately represent spatial connections across far-flung face areas. Transformer-based models are able to better capture local and global face signals by incorporating attention processes at several feature levels. For strong FER in challenging, real-world situations, such designs are therefore becoming more and more popular.

FER systems have advanced significantly as outlined above. However, there are still a number of research gaps that prevent their practical applications. Due to variances in demography, lighting, and face features, many current models have trouble generalizing across datasets; they perform well on certain datasets but fall short when evaluated on unseen data \cite{zhang2023dual, wang2024qcs,huang2023study}. Conventional CNN-based FER models are limited in their capacity to adapt to a variety of inputs because they rely on set feature extractors, which might not be the best for all facial expressions \cite{uniyal2024analyzing,huang2023study}. Furthermore, it is challenging to capture both global face structures and fine-grained expression features since the majority of architectures do not integrate multi-scale feature extraction \cite{uniyal2024analyzing,wang2024qcs,huang2023study}. Managing occlusions, light fluctuation, and non-frontal head poses—all of which impair model performance, this is another crucial issue in real-world FER.

Dataset imbalances also make it difficult to classify underrepresented emotions like disgust, fear, and contempt, which restricts the model's capacity to confidently identify uncommon expressions which is evident in \cite{bohi2024novel,zhang2023dual,wang2024qcs}. Low-resolution grayscale images and mislabeling problems in the popular dataset FER-2013 pose additional difficulties that can have a big influence on model learning and generalization. By utilizing a Mixture of Experts (MoE) framework for adaptive feature selection, incorporating multiple CNN-based extractors for multi-scale learning, and improving generalization across datasets by combining deep feature learning and ensemble-based decision-making, ExpressNet-MoE directly addresses these limitations. ExpressNet-MoE is a very flexible and scalable solution for real-world FER applications since it enhances robustness against occlusions, illumination shifts, and dataset bias by dynamically choosing expert networks based on input data.

\section{DATASETS}
\label{sec:dataset}
We have used three benchmark datasets which are AffectNet, RAF-DB, and FER-2013 to train our model. Each of the aforementioned datasets have distinct qualities that will help improve robustness and generalizability of the model. AffectNet is perfect for learning complicated emotional variations since it offers a large collection of face expressions that have been taken in real-world situations. RAF-DB's extensive collection of extremely diverse photos with well-documented annotations will help the model to test its adaptability and generalizability. Furthermore, the grayscale photographs in FER-2013 were taken in a variety of settings to help the model adapt to variations in lighting and facial expressions.

\begin{figure*}[tph]
    \includegraphics[scale=0.17]{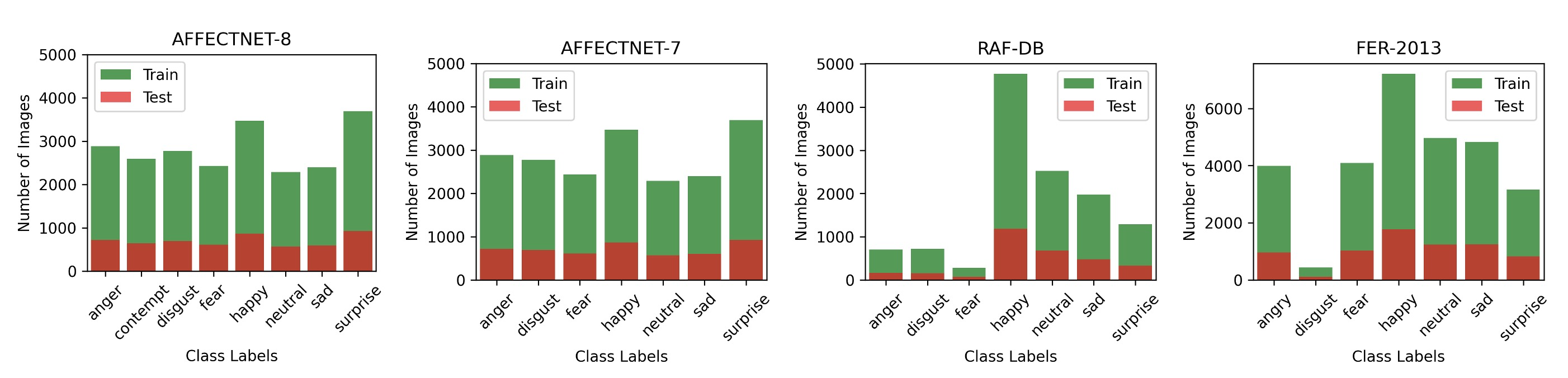}
    \caption{Training vs Test Splits for the datasets. \label{fig:datasets}}
\end{figure*}

In order to balance computational efficiency, and memory constraints, we employed subsets from AffectNet and RAF-DB. 
Fig. \ref{fig:datasets} illustrates the total number of images for every dataset per emotion class along with their training and testing splits along with the class distributions.  %is compiled in Table \ref{tab:datasets}. 
%Furthermore, the distribution for each dataset is plotted in Fig. \ref{fig:datasets}. 
AffectNet's 8-class classification is denoted by $\text{AffectNet}_8$, whereas AffectNet's 7-class classification is denoted by $\text{AffectNet}_7$. We increased the model's adaptability by utilizing these datasets making the model more robust. 

%\begin{table*}[tph]
%    \caption{Datasets and their total images, train and test splits. \label{tab:datasets}}
%    \begin{tabular}{l|ccc|cccc|cccc}
%    \toprule
%    Split & \multicolumn{3}{c|}{Total} & \multicolumn{4}{c|}{Train} & \multicolumn{4}{c}{Test}\\\midrule
%\footnotesize{Emotion} & \footnotesize{Affectnet}	& \footnotesize{RAF-DB} & \footnotesize{FER-2013} & \footnotesize{$\text{Affectnet}_8$} & \footnotesize{$\text{Affectnet}_7$} & \footnotesize{RAF-DB} & \footnotesize{FER-2013} & \footnotesize{$\text{Affectnet}_8$} & \footnotesize{$\text{Affectnet}_7$} & \footnotesize{RAF-DB} & \footnotesize{FER-2013}\\\midrule
%Surprise & 4616 & 1619 & 4002 & 3693 & 3693 & 1290 & 3171 & 923 & 923 & 329 & 831\\
%Happy & 4336 & 5957 & 8989 & 3469 & 3469 & 4772 & 7215 & 867 & 867 & 1185 & 1774\\
%Anger & 3608 & 867 & 4953 & 2886 & 2886 & 705 & 3995 & 722 & 722 & 162 & 958\\
%Disgust & 3472 & 877 & 547 & 2778 & 2777 & 717 & 436 & 694 & 695 & 160 & 111\\
%Neutral & 2861 & 3204 & 6198 & 2289 & 2289 & 2524 & 4965 & 572 & 572 & 680 & 1233\\
%Fear & 3043 & 355 & 5121 & 2434 & 2434 & 281 & 4097 & 609 & 609 & 74 & 1024\\
%Sad & 2995 & 2460 & 6077 & 2396 & 2396 & 1982 & 4830 & 599 & 599 & 478 & 1247\\
%Contempt & 3244 & - & - & 2595 & - & - & - & 649 & - & - & -\\\midrule
%Total & 28175 & 15339 & 35887 & 22540 & 19944 & 12271 & 28709 & 5635 & 4987 & 3068 & 7178\\\bottomrule
%    \end{tabular}
%\end{table*}

\subsection{AffectNet}
AffectNet \cite{mollahosseini2017affectnet} is one of the biggest datasets for facial expression identification. It contains over a million photos and about 440,000 samples have annotations. We have chosen as subset of this dataset with 28,175 photos from eight different emotional categories from the mentioned dataset for this study which are surprise, happy, anger, disgust, neutral, fear, sad, and contempt.
%, from Kaggle [21]. 
These photos are collected from online sources and are taken from actual real-life situations and contain a variety of facial expressions. Because of its vast size and computational constraints, a subset of this dataset is generally used by many researchers in their works \cite{bhati2025generalized,uniyal2024analyzing,savchenko2022classifying}.

For training, we have used 22,540 images, and 5,635 were set aside for testing. The selected dataset includes a wide range of age groups, ethnicity, and lighting conditions, making it ideal for improving generalization and the model’s adaptability. To make a 7-class classification on the AffectNet dataset, we remove the Contempt category and made a training set of 19,944 images and a testing set of 4,987. Although AffectNet offers a wide variety of emotions, the distribution of classes is imbalanced compared to the original which is heavily imbalanced, with Happy and Surprise being more represented than Neutral and Fear.

Because of its versatility, AffectNet is a great tool for training models to function well in multiple settings. The excellent careful annotations in the dataset ensure that the model learns from precise labels and aid in improving emotion classification. The dataset is a popular option for emotion detection despite its class imbalance.

\subsection{RAF-DB}
Another popular dataset for facial emotion detection is RAF-DB \cite{li2017reliable,li2019reliable}.
%we got this from Kaggle as well [23]. This website wants us to cite the two above: http://www.whdeng.cn/RAF/model1.html
It consists of 15,339 single class photos tagged with the seven main emotions classes that are surprise, fear, happiness, anger, disgust, neutrality, and sadness. It contains pictures of people of all ages, genders, and cultural backgrounds. It stands out for its diversity. This variation in the dataset makes it effective for training models that must function well across various populations. 

We used 3,068 photos for testing and 12,271 images for training from the RAF-DB dataset. The photographs in this dataset like AffectNet were also taken in real-world settings and are downloaded from the internet. It contains variations in head positions, backgrounds, and lighting. Building a strong model that can identify emotions outside of controlled environments requires an understanding of these real-world difficulties. There is a noticeable class disparity in RAF-DB, as happy and neutral classes are significantly more prevalent than fear, anger and disgust.

We proved the model's generalizability by identifying nuanced emotional features by using RAF-DB dataset. The dataset complements the AffectNet and FER-2013 datasets, as the dataset guarantees a more comprehensive portrayal of emotions. The reliability of the emotion classification model is further increased by the high-quality training data provided by its well-annotated images.

\subsection{FER-2013}
FER-2013 \cite{goodfellow2013challenges}, which was first presented in the ICML 2013 challenge, is a well-known dataset in facial expression recognition. There are 35,887 grayscale photographs in the dataset, each with a resolution of 48 by 48 pixels. It contains 7 different emotion classes which are anger, disgust, fear, happiness, neutrality, sadness, and surprise.
%Citation of FER-2013 is fixed.

Although FER-2013 is widely used, it has a number of drawbacks. The first one is mislabeling \cite{barsoum2016training,mazen2021real}, wherein some images are mistakenly categorized as belonging to the wrong class of emotion. This may impact the model’s performance as it introduces noise during training. Furthermore, the low-resolution photos in the dataset makes it more difficult for models to extract fine-grained facial traits. On top of that, emotions like happy and neutral have far more samples than other classes like disgust and surprise, indicating a class imbalance. The model may become less sensitive to underrepresented emotions as a result of this imbalance, which could result in biased  predictions. These drawbacks are somewhat offset by its size and variety of real-world examples in the classes with more samples.
%Citation to FER+ is changed above.
For this study, we utilized 28,709 images for training and 7,178 for testing. 
%However, given its limitations, FER-2013 might not be enough on its own for high-accuracy emotion identification.

\section{PROPOSED METHODOLOGY}
\label{sec:proposed_method}
The proposed methodology involves training the model on three facial expression recognition datasets: AffectNet, RAF-DB, and FER-2013 (individually, to test generalizability). Two variations of AffectNet were taken into consideration: one with seven emotion classes with the "contempt" category eliminated - $\text{AffectNet}_7$, and another with all the eight classes kept - $\text{AffectNet}_8$. There are seven emotion classes in RAF-DB as in $\text{AffectNet}_7$. To maintain compatibility with other datasets for the model's input, the 
%seven classes in 
grayscale images in 
FER-2013 were
%must be 
transformed 
%converted 
into three channels. To guarantee an equitable distribution of emotion categories in both training and testing sets, a stratified train-test split is used for all datasets. This keeps the model's performance from being distorted by data imbalance as much as possible. We will be using the same test split for evaluating our model as well as comparing our proposed model with the state-of-the-art methods. %validation since we are not performing any hyperparameter fine-tuning using the validation set. Therefore, the model never really “sees” the validation set and is suitable for testing as well.

\begin{figure}[!ht]
    \centering
    \includegraphics[scale=0.5]{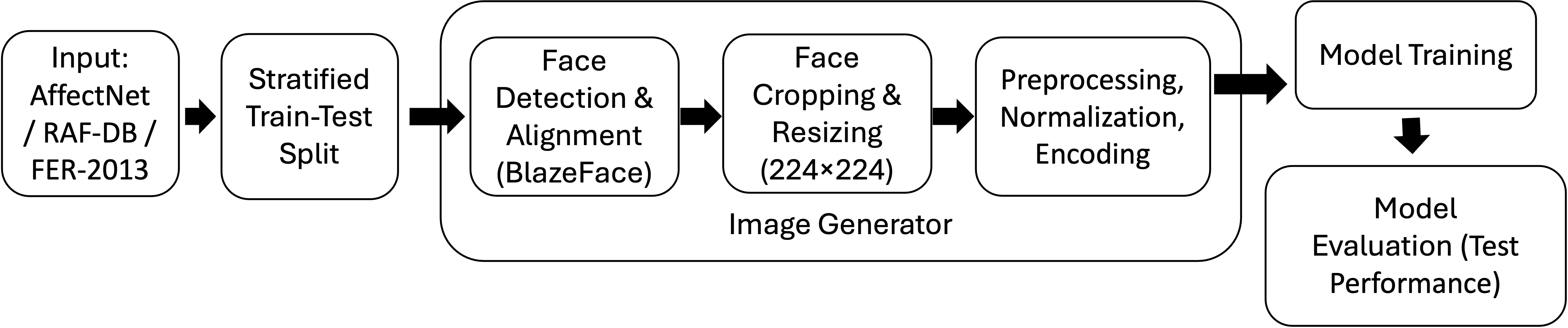}
    \caption{Proposed methodology\label{fig:proposed_methodology}}
\end{figure}

The BlazeFace model \cite{bazarevsky2019blazeface} from MediaPipe, which effectively identifies facial areas in photos, is used for face identification and alignment. It is used by many researchers for preprocessing facial images \cite{hossen2023dataset,soudy2024deepfake,bayar2022novel}. The identified faces are cropped after the relative bounding-box coordinates are calculated and gently adjusted for robustness. The original image is scaled to $224\times 224$ pixels if no face is found. An image generator that dynamically loads photos, does preprocessing (using BlazeFace), normalizes pixel values, and one-hot encodes emotion labels is included into the data pipeline. To maximize training efficiency and avoid overfitting, the model is trained for 15 epochs with a batch size of 32 using methods such as checkpointing, learning rate adjustments, and early stopping. This methodology ensures effective learning across diverse datasets and also improving the model’s generalization for facial expression recognition. Fig. \ref{fig:proposed_methodology}. shows the complete methodology.

Accuracy, precision, recall, and $F_1$-score are used to evaluate facial expression recognition algorithms.
% DB - adding content
Accuracy measures the overall proportion of correct predictions, giving a general sense of how well the model performs. Precision focuses on how many of the predicted expressions were actually correct. Recall, on the other hand, assesses how many of the true expressions the model was able to identify. The F1-score combines precision and recall into a single value, and provides a balanced measure which is especially useful when a dataset is imbalanced.
% DB - Content end.
These metrics provide a thorough evaluation of the model's performance. 
%Although accuracy measures how accurate predictions are overall, it might be deceptive for datasets that are unbalanced, it is calculated using Equation \ref{eqn:Acc}. True Positive (TP) is the correct prediction of a positive outcome, True Negative (TN) is when a prediction correctly identifies that a condition or characteristic is absent when it actually is. When a model predicts a positive class for an instance that really belongs to a negative class, this is known as a False Positive (FP). When a model predicts a negative class for an instance that really belongs to a positive class, this is known as a False Negative (FN). However, recall evaluates how effectively the model detects real positives (Equation \ref{eqn:rec}), whereas precision shows how many predicted positive cases are actually true (Equation \ref{eqn:prec}). When class distribution is unequal, the $F_1$-score provides a more dependable statistic by striking a balance between precision and recall, which is calculated using Equation \ref{eqn:f1}. 

%\begin{align}
%    \text{Accuracy} &=& \dfrac{TP+TN}{TP+TN+FP+FN}\label{eqn:Acc}\\
%    \text{Recall} &=& \dfrac{TP}{TP+FN}\label{eqn:rec}\\
%    \text{Precision} &=& \dfrac{TP}{TP+FP} \label{eqn:prec}\\
%    F_1 &=& \dfrac{2\cdot \text{Precision}\cdot \text{Recall}}{\text{Precision}+ \text{Recall}}\label{eqn:f1}
%\end{align}

%When combined, these metrics provide a thorough evaluation of the model's performance.

%\subsection{MODEL ARCHITECTURE}
\section{MODEL ARCHITECTURE}
\label{sec:model_architecture}
% Hyperparameter details
A number of significant hyperparameters that control the component structure and training behavior incorporated into the proposed architecture are listed in this section. These include parameters associated with the Mixture of Experts (MoE) module, such as the number of experts and gating methods, as well as kernel sizes, filter counts, dropout rates, activation functions, and normalization strategies within the CNN-based  feature extractors. It also includes the hyperparameters used in training such as learning rate, batch size, optimizer, and loss function. The sections that follow offer a thorough description of these hyperparameters. We tried several different hyperparameters for the model including different filter sizes, activation functions, depth of CNN layers, etc. and finally selected parameters that worked best and still provided a reasonable model size.

\subsection{CNN Feature Extractor 1}
\begin{figure}[!ht]
    \centering
    \includegraphics[scale=0.65]{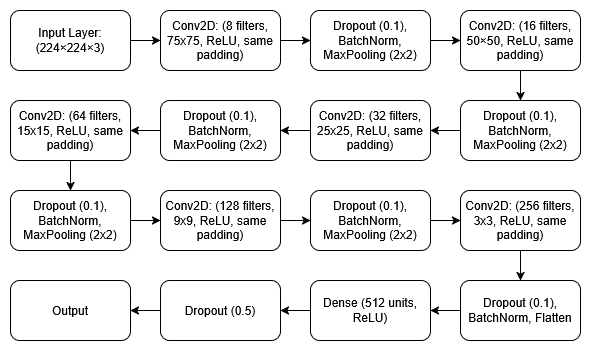}
    \caption{CNN Feature Extractor 1 (CNNFE1)\label{fig:CNNFE1}}
\end{figure}

CNN Feature Extractor 1 or CNNFE1 (Fig. \ref{fig:CNNFE1}) is a Deep Convolution Neural Network (D-CNN) designed to extract hierarchical spatial information from a $224\times 224\times 3$ input images.

%\begin{figure}[!ht]
%    \centering
%    \includegraphics[scale=0.32]{figs/CNNFE1.png}
%    \caption{CNN Feature Extractor 1 (CNNFE1)\label{fig:CNNFE1}}
%\end{figure}

To preserve spatial dimensions, the network begins with a $75\times 75$ convolution layer with 8 filters, ReLU activation, and ``same'' padding. The convolution operation is defined in Equation~\ref{eqn:conv}, where $Y(i,j)$ is the output feature at position $(i,j)$, $X(i+m,j+n)$ is the input or previous layer's output, and $K(m,n)$ is the filter kernel.

\begin{align}
    Y(i,j) = \sum_m\sum_n X(i+m,j+n)*K(m,n)\label{eqn:conv}
\end{align}

%Where $Y(i,j)$ is going to be the output feature map at position $(i,j)$. $X(i+m,j+n)$ represents the input image or the previous layer’s feature map. $K(m,n)$ is the convolution kernel or filter. The summation slides over the input and across the filter size $m*n$.

%The ReLU formula is given in Equation \ref{eqn:relu}. ReLU is an activation function that is commonly used in Neural Networks, mostly in the middle layers. It takes an input x and calculates the max between x and 0, which means if x is negative, 0 is returned and if x is positive, x is returned. ReLU introduces non-linearity and maintains computational efficiency. It also mitigates the vanishing gradients (unlike functions like sigmoid or tanh) problem, which allows deeper neural networks to train more efficiently.
% Addded to reduce words
The ReLU formula is given in Equation \ref{eqn:relu}. ReLU (Rectified Linear Unit) is a widely used activation function that outputs the input (x) if positive, or zero otherwise. It introduces non-linearity, it is computationally efficient, and helps prevent vanishing gradients. This enables deeper networks to train effectively.

\begin{align}
    f(x) = \text{max}(0,x)\label{eqn:relu}
\end{align}

%After that, a dropout layer with a rate of 0.1 randomly deactivates neurons to mitigate overfitting. After that, batch normalization is used to speed up convergence and stabilize training. Batch Normalization is done using a series of steps. Given in Equations \ref{eqn:mB},\ref{eqn:s2_B}, \ref{eqn:c_i}, and \ref{eqn:y_i}.
A dropout layer (rate = 0.1) follows to reduce overfitting, then batch normalization is applied to stabilize and accelerate training. Given input $x_i$ in a mini-batch of size $m$, batch normalization proceeds as follows:

\begin{align}
    m_B &= \dfrac{1}{m}\sum_{i=1}^m x_i \label{eqn:mB} \\
    s^2_B &= \dfrac{1}{m}\sum_{i=1}^m (x_i - m_B)^2 \label{eqn:s2_B} \\
    c_i &= \dfrac{x_i - m_B}{\sqrt{s^2_B + \epsilon}} \label{eqn:c_i} \\
    y_i &= \gamma c_i + \beta \label{eqn:y_i}
\end{align}
%\begin{align}
%    m_B = \dfrac{1}{m}\sum_{i=1}^m x_i \label{eqn:mB}
%\end{align}

Equations~\ref{eqn:mB} and \ref{eqn:s2_B} compute the batch mean $m_B$ and variance $s^2_B$ for input $x_i$, where $m$ is the batch size.
Equation~\ref{eqn:c_i} normalizes the input using the batch statistics, with $\epsilon$ preventing division by zero.
Here, $\gamma$ and $\beta$ are learnable scale and shift parameters used in batch normalization.
%Here, $\epsilon$ prevents division by zero, and $\gamma$, $\beta$ are learnable parameters. 
A $2\times 2$ max-pooling layer then reduces spatial dimensions and improves efficiency. The max-pooling operation is defined in Equation~\ref{eqn:y_i_j}.

\begin{align}
    Y(i,j) = \underset{m,n}{\text{max}}\, X(i+m, j+n) \label{eqn:y_i_j}
\end{align}

Following the first convolution layer, the feature extraction process continues with progressively smaller kernels and more filters: $50\times 50$ (16), $25\times 25$  (32), $15\times 15$ (64), $9\times 9$ (128), and $3\times3$ (256), each followed by dropout, batch normalization, and pooling (except the last). The output is flattened and passed to a dense layer with 512 neurons (ReLU).
%with a $50\times 50$ convolution layer with 16 filters, maintaining the ReLU activation and batch normalization structure, followed by another $2\times 2$ max-pooling operation. This pattern repeats with increasing filter sizes—32 filters with a $25\times 25$ kernel, 64 filters with a $15\times 15$ kernel, 128 filters with a $9\times 9$ kernel, and finally 256 filters with a $3\times 3$ kernel. Each convolution layer is followed by dropout (0.1), batch normalization, and max pooling (except the last). Before going through a deep (Dense) layer of 512 neurons with ReLU activation, the final feature maps are flattened into a 1D vector. 
The dense layer (i.e., fully connected layer) calculates its outputs ($y$) using its definition listed in Equation \ref{eqn:y}. 

\begin{align}
    y = f(Wx+b)\label{eqn:y}
\end{align}

$W$ is the weight matrix of the layer and $b$ is the bias vector for the current layer. $f$ is the activation function used (in this case, ReLU).
% DB - unnecessary
%Where $W_{ij}$ is representing the weight connecting the $i$-th neuron in the previous layer to the $j$-th neuron in the current layer. $h$ is the input vector to the dense layer and $b$ is the bias vector for the current layer. $f$ is the activation function used (in this case, ReLU).

This is finally followed by dropout (0.5) to prevent overfitting. The model captures both low and high-level spatial features efficiently.

%Before the recovered features are sent to the next network component, a last dropout layer with a rate of 0.5 is performed to avoid overfitting. The model should be able to gradually capture both low-level and high-level spatial patterns.

\subsection{CNN Feature Extractor 2}
\begin{figure}[!ht]
    \centering
    \includegraphics[scale=0.65]{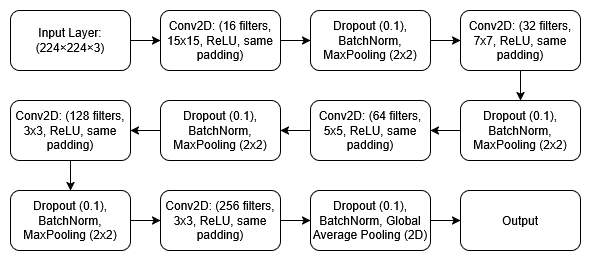}
    \caption{CNN Feature Extractor 2 (CNNFE2)\label{fig:CNNFE2}}
\end{figure}

CNN Feature Extractor 2 or CNNFE2 as shown in Fig. \ref{fig:CNNFE2}. is also a CNN architecture designed to extract hierarchical spatial features while maintaining computational efficiency.

%\begin{figure}[!ht]
%    \centering
%    \includegraphics[scale=0.32]{figs/CNNFE2.png}
%    \caption{CNN Feature Extractor 2 (CNNFE2)\label{fig:CNNFE2}}
%\end{figure}

It starts with a 15×15 convolution layer (16 filters, ReLU, same padding), followed by dropout (0.1), batch normalization, and 2×2 max pooling. This structure repeats with 32 filters (7×7), 64 filters (5×5), 128 filters (3×3), and finally 256 filters (3×3), each layer followed by batch normalization, ReLU, and pooling. Batch normalization is used to stabilize the training process and the dropout layers are applied to reduce overfitting. Instead of flattening (like in CNNFE1), a Global Average Pooling (GAP) layer reduces each feature map to a single value, creating a compact and efficient feature vector. GAP for a given feature map $X$ of size $H \times W$ (height $\times$ width) is computed by the equation given in Equation \ref{eqn:Y_from_cnnfe2}.
 % DB - removed to ma ke concise
 %In order to maintain spatial dimensions, it begins with a $15\times 15$ convolution layer with 16 filters that uses ReLU activation and identical padding. Batch normalization is used to stabilize the training process after a dropout layer with a rate of 0.1 is applied to reduce overfitting. While maintaining key features, a $2\times 2$ max-pooling layer lowers spatial resolution. A $7\times 7$ convolution layer with 32 filters is introduced in the following step, which again uses batch normalization and ReLU activation before max-pooling is applied. Then, using the same organized method of dropout, batch normalization, and pooling, the network deepens with a $5\times 5$ convolution layer with 64 filters. A $3\times 3$ convolution layer with 128 filters is used to further refine feature representation as the complexity of the extracted features rises. Lastly, a final convolution layer, which uses 256 filters with a $3\times 3$ kernel. By averaging each feature map into a single value, a global average pooling (GAP) layer is utilized in place of flattening the feature maps, therefore lowering spatial dimensions and producing a compact feature vector. GAP for a given feature map $X$ of size $H \times W$ (height $\times$ width) is computed by the equation given in Equation \ref{eqn:Y_from_cnnfe2}.

 \begin{align}
     Y = \dfrac{1}{H\cdot W}\sum_{i=1}^H\sum_{i=1}^W X(i,j) \label{eqn:Y_from_cnnfe2}
 \end{align}

 $X(i,j)$ is the value of the feature map at position $(i,j)$ and $Y$ is the scalar output representing the average value of the entire feature map.

\subsection{ResNet-50 Model}
\begin{figure}[!ht]
    \centering
    \includegraphics[scale=0.65]{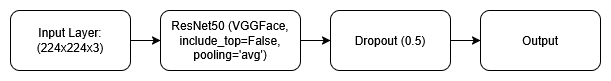}
    \caption{ResNet-50 Architecture\label{fig:resnet50}}
\end{figure}

A pre-trained ResNet-50 model trained on the VGGFace2 dataset \cite{cao2018vggface2} originally tasked for face recognition was leveraged here to acquire the high-level face features. It ensures deep feature learning by processing $224\times 224 \times 3$ images through the residual blocks of ResNet50. We have used \texttt{include\_top=false} that turns the model into a pure feature extractor by removing the classification levels. In order to avoid overfitting, a 0.5 dropout layer is used after global average pooling which condenses feature maps into a compact vector. Transfer learning is used in this architecture to provide reliable face expression recognition. This sub-model architecture is shown in Fig. \ref{fig:resnet50}.

%\begin{figure}[!ht]
%    \centering
%    \includegraphics[scale=0.32]{figs/resnet50.png}
%    \caption{ResNet-50 Architecture\label{fig:resnet50}}
%\end{figure}

\subsection{Mixture of Experts}
\begin{figure}[!ht]
    \centering
    \includegraphics[scale=0.65]{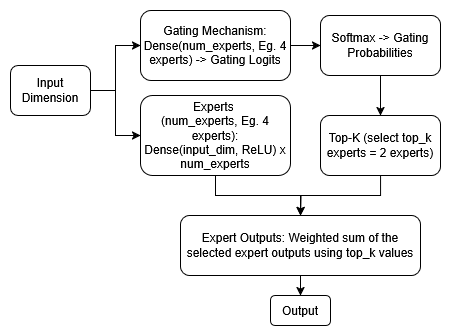}
    \caption{Mixture of Experts (MoE)\label{fig:MoE}}
\end{figure}

Our model's Mixture of Experts (MoE) module enhances the performance by dynamically choosing the most pertinent expert or experts for each input by utilizing a variety of expert networks. The structure of MoE is given in Fig. \ref{fig:MoE}.

%\begin{figure}[!ht]
%    \centering
%    \includegraphics[scale=0.32]{figs/MoE.png}
%    \caption{Mixture of Experts (MoE)\label{fig:MoE}}
%\end{figure}

The network takes a vector of size \texttt{input\_dim} as input and starts with a dense input layer. After that, this input is run in parallel through a number of expert models (denoted by \texttt{num\_experts}, e.g., 4 experts), each of which has a dense layer with ReLU activation to enable it to capture distinct data features. To regulate the information flow, a gating network is implemented, with \texttt{num\_experts} logits generated by its own dense layer. A Softmax activation is used to convert these logits into a probability distribution over the experts, allowing the gating mechanism to choose the final experts. The softmax is defined Equation \ref{eqn:softmax}. 

\begin{align}
    %\text{softmax}_i = \dfrac{e^{z_i}}{\sum_{j=1}^K e^{z_j}}\label{eqn:softmax}
    \text{softmax}(z_i) = \frac{e^{z_i}}{\sum_j e^{z_j}}, \label{eqn:softmax}
\end{align}

Where, $z_i$ is the raw output (logit) for class $i$ and $e^{z_i}$ is the exponential of the logit $z_i$. The denominator is the summation of the exponentials of all logits, this ensures that the output probabilities sum to 1.

After stacking the expert outputs into a tensor, the top-$k$ function from TensorFlow is used to choose the top-$k$ experts (2 experts, default) based on the gating probabilities. Each expert's contribution is scaled by its corresponding gating probability, and the final result is calculated as a weighted sum of the top expert outputs. This makes it possible for the model to concentrate on the experts who are most pertinent to each input, facilitating more specialized and effective decision-making. The original input is then transformed into the weighted combination of the top expert outputs by wrapping the MoE layer in a Tensorflow/Keras Model. The model's ability to adaptively select experts through the MoE architecture improves performance on tasks involving complex decision-making.

\subsection{Our Final Model -- ExpressNet-MoE}
\begin{figure*}[!ht]
    \centering
    \includegraphics[scale=0.41]{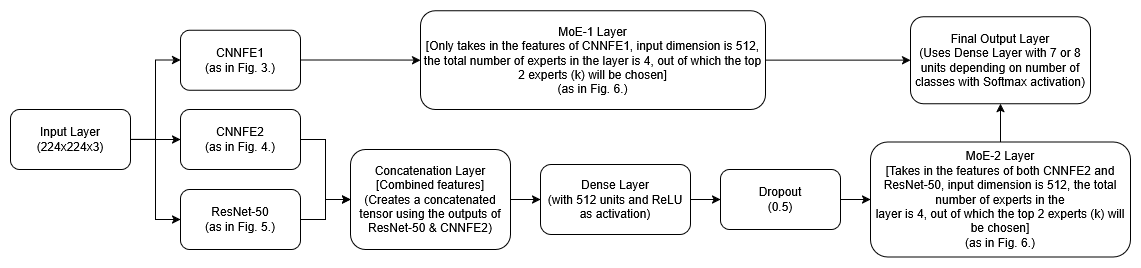}
    \caption{ExpressNet-MoE Architecture\label{fig:ExpressNet_MoE}}
\end{figure*}

The CNNFE1, CNNFE2, ResNet-50 and MoE modules are combined in the final ExpressNet-MoE Model (Fig. \ref{fig:ExpressNet_MoE}) to produce a robust architecture for emotion recognition. 

%\begin{figure}[!ht]
%    \centering
%    \includegraphics[scale=0.32]{figs/ExpressNet-MoE.png}
%    \caption{ExpressNet-MoE Architecture\label{fig:ExpressNet_MoE}}
%\end{figure}

The model starts with a Keras Input layer that can handle $224\times 224$ RGB images. In order to capture varying amounts of picture information, three different CNN-based feature extractors are used: CNNFE1, CNNFE2, and ResNet50. Sequentially putting the input through these models yields their results. Following the concatenation of the CNNFE2 and ResNet50 features, a dense layer with 512 units and ReLU activation is applied to the combined features. A Dropout layer with a regularization rate of 0.5 is then applied.

The two sets of features are subjected to the MoE mechanism: a MoE layer specifically designed for CNN-based  features processes the features from CNNFE1, and a second MoE layer also designed for the CNN features processes the dense layer output (from CNNFE2 and ResNet50). The high-level extracted representations of the input image are represented by a composite feature vector created by concatenating the outputs from the two MoE layers which we will call the final features.

To predict one of the emotions in input images, an output layer with softmax activation is added at the end taking the output as an input from the final features. To enhance generalization, the model is assembled using the Adam optimizer and categorical cross-entropy loss with label smoothing. The chosen loss function, categorical Cross Entropy (CCE) is calculated using Equation \ref{eqn:CCE}. 
In the equation, $K$ is the number of classes, $y_i$ is the true label and $\hat{y}_i$ 
is the predicted label. 
$\log(\hat{y}_i)$ is the natural logarithm of the predicted probability of the class $i$.

\begin{align}
    \text{CCE}(y,\hat{y}) = -\sum_{i=1}^K y_i*\log \hat{y}_i\label{eqn:CCE}
\end{align}

We ran our experiments in a Tensorflow 2.0 compute framework configured with four NVidia Titan Xp GPU co-processors with mirrored strategy enabled. Values of the hyperparameters were chosen by heuristics and are listed in Table \ref{tab:hyperparams}. %We used a batch size of 32 for training and validation as it keeps memory usage low enough to fit within the constraints of most modern GPUs for efficient training. We also used early stopping which avoids overfitting by stopping training if validation loss does not improve after five consecutive epochs, while a learning rate scheduler lowers the learning rate by 5\% each epoch. In order to save the optimal model based on validation accuracy, model checkpoints are enabled. TensorFlow's Mirrored Strategy enables multi-GPU training (e.g., in our experiments it was 4) to scale this architecture for huge datasets, guaranteeing effective training across numerous devices.

\begin{table}[!ht]
    \centering
    \caption{List of Hyperparameter Values\label{tab:hyperparams}}
    \begin{tabular}{lc}
    \toprule
    \textbf{Hyperparameter} & \textbf{Value}\\
    \midrule
    Neural Network type & Hybrid CNN-based Mixture of Experts\\\midrule
    Classification type & Multi-class (7, and 8 classes)\\\midrule
    Batch size & 32\\\midrule
    Dropout rate & Variable (0.1-0.5)\\\midrule
    Number of epochs & 15\\\midrule
    \multirow{3}{*}{Optimizer} & Adam\\
    & Learning rate, $\alpha=10^{-4}$\\
    & $\beta_1=0.9, \beta_2=0.999$\\\midrule
    Loss function & Categorical Cross-entropy\\\bottomrule
    \end{tabular}
\end{table}

\section{RESULTS}
\label{sec:results}
Table \ref{tab:eval_all_dataset} lists the classification report of our proposed ExpressNet-MoE model on all of the four datasets in terms of precision, recall, $F_1$, and per-class accuracy ($p$-Acc), while Table \ref{tab:test_acc_db} lists the overall accuracy, macro averaged precision, recall and $F_1$-score of our model at the evaluation.

%The classification report for all the datasets for testing by the model is given in Table  and the final testing accuracies for the datasets are given in Table .

\begin{table}[!ht]
    \centering
    \caption{Classification Report of proposed ExpressNet-MoE model on all the test-sets from $\text{AffectNet}_7$ (AN-7), $\text{AffectNet}_8$ (AN-8), RAF-DB (RAF), FER-2013 (FER). \label{tab:eval_all_dataset}}
    \begin{tabular}{lcccccc}
    \toprule
    Emotion & Dataset & Precision & Recall & $F_1$ & Support & $p$-Acc\\
    \midrule
\multirow{4}{*}{Anger} & AN-7 & 0.71 & 0.79 & 0.75 & 722 & 92.38\\
 & AN-8 & 0.83 & 0.55 & 0.66 & 722 & 92.80\\
 & RAF & 0.71 & 0.87 & 0.78 & 162 & \textbf{97.43}\\
 & FER & 0.61 & 0.58 & 0.59 & 958 & 89.40\\\midrule
\multirow{4}{*}{Disgust} & AN-7 & 0.79 & 0.52 & 0.63 & 695 & 91.46\\
  & AN-8 & 0.68 & 0.72 & 0.70 & 694 & 92.48\\
  & RAF & 0.62 & 0.60 & 0.61 & 160 & 96.02\\
  & FER & 0.75 & 0.36 & 0.49 & 111 & \textbf{98.83}\\\midrule
\multirow{4}{*}{Fear} & AN-7 & 0.70 & 0.76 & 0.73 & 609 & 93.18\\
  & AN-8 & 0.90 & 0.59 & 0.71 & 609 & 94.85\\
  & RAF & 0.65 & 0.62 & 0.63 & 74 & \textbf{98.27}\\
  & FER & 0.65 & 0.29 & 0.40 & 1024 & 87.66\\\midrule
\multirow{4}{*}{Happy} & AN-7 & 0.80 & 0.97 & 0.88 & 867 & 95.31\\
  & AN-8 & 0.89 & 0.84 & 0.86 & 867 & \textbf{95.94}\\
  & RAF & 0.91 & 0.95 & 0.93 & 1185 & 94.43\\
  & FER & 0.88 & 0.84 & 0.86 & 1774 & 93.22\\\midrule
\multirow{4}{*}{Neutral} & AN-7 & 0.70 & 0.72 & 0.71 & 572 & \textbf{93.18}\\
  & AN-8 & 0.55 & 0.73 & 0.62 & 572 & 91.14\\
  & RAF & 0.88 & 0.73 & 0.80 & 680 & 91.75\\
  & FER & 0.50 & 0.75 & 0.60 & 1233 & 82.84\\\midrule
\multirow{4}{*}{Sad} & AN-7 & 0.76 & 0.62 & 0.68 & 599 & 93.06\\
  & AN-8 & 0.62 & 0.75 & 0.68 & 599 & 92.56\\
  & RAF & 0.77 & 0.88 & 0.82 & 478 & \textbf{93.97}\\
  & FER & 0.50 & 0.64 & 0.56 & 1247 & 82.77\\\midrule
\multirow{4}{*}{Surprise} & AN-7 & 0.75 & 0.76 & 0.76 & 923 & 90.98\\
  & AN-8 & 0.74 & 0.80 & 0.77 & 923 & 92.00\\
  & RAF & 0.87 & 0.82 & 0.84 & 329 & \textbf{96.71}\\
  & FER & 0.85 & 0.64 & 0.73 & 831 & 94.61\\\midrule
Contempt & AN-8 & 0.69 & 0.76 & 0.72 & 649 & \textbf{93.33}\\\bottomrule
    \end{tabular}
\end{table}

\begin{table}[!ht]
    \centering
    \caption{Overall Evaluation of ExpressNet-MoE model on the test-sets in terms of accuracy and macro averaged Precision, Recall and $F_1$-score. \label{tab:test_acc_db}}
    \begin{tabular}{lcccc}
    \toprule
    Dataset & $\text{AffectNet}_7$ & $\text{AffectNet}_8$ & RAF-DB & FER-2013 \\
    \midrule
    Accuracy        & 74.77\% & 72.55\%  & 84.29\%	& 64.66\%\\
    Macro-Prec. & 74.56\% & 73.69\%  & 77.28\%  & 67.86\%\\
    Macro-Rec.    & 73.58\% & 71.83\%  & 78.00\%  & 58.62\% \\
    Macro-$F_1$     & 73.41\% & 71.75\%  & 77.34\%  & 6058\%\\\bottomrule
    \end{tabular}
\end{table}

The model's performance varies on different datasets, which reflects issues specific to each dataset. Happiness regularly has the highest $F_1$-scores, showing a good recognition of positive emotions, especially in RAF-DB (0.93) and FER-2013 (0.86). In FER-2013, anger and fear had lower $F_1$-scores (0.59 and 0.40, respectively), due to class imbalance and incorrect classification. In contrast to AN-8 (0.70), disgust performs badly in FER-2013 ($F_1$ = 0.49), which is directly correlated to having less training and testing data. In FER-2013, neutral expressions have a high recall (0.75) but low precision (0.50), which compromises the reliability of classification.   Overall, our model shows consistent performance on all of RAF-DB emotion classes (84.29\%) (Table \ref{tab:test_acc_db}), yet it shows similar on prediction accuracies on the two versions of AffectNet datasets: $\text{AffectNet}_7$ (74.77\%) and $\text{AffectNet}_8$ (72.55\%).
%have close enough accuracies with $\text{AffectNet}_7$ having a higher accuracy (74.77\%) than $\text{AffectNet}_8$ (72.55\%). Out of the four datasets RAF-DB reports the highest testing accuracy (84.29\%).

The generalized trend noticed during the training on all datasets is that validation accuracy typically followed an increasing trend with some fluctuations, indicating possible overfitting, training accuracy demonstrated consistent improvements. Our check-points made sure only the model with the highest validation accuracy was saved to mitigate the overfitting. Adjusting the learning rate facilitated smooth convergence, while early stopping avoided needless training. 

%\begin{table}[!ht]
%    \centering
%    \caption{Macro-averaged evaluation of ExpressNet-MoE\label{tab:macro-eval}}
%    \begin{tabular}{lcccc}
%    \toprule
%    \textbf{Dataset} & \textbf{Accuracy} & \textbf{Precision} & \textbf{Recall} & \textbf{$F_1$-score}\\
%    \midrule
%    AffectNet-7 (AN-7) & 0.75 & 0.75 & 0.74 & 0.73\\
%    AffectNet-8 (AN-8) & 0.73 & 0.74 & 0.72 & 0.72\\
%    RAF-DB (RAF) & 0.84 & 0.77 & 0.78 & 0.77\\
%    FER-2013 (FER) & 0.65 & 0.68 & 0.59 & 0.61\\
%    \bottomrule
%    \end{tabular}
%\end{table}

\subsection{$\text{AffectNet}_7$ Results}
%Fig. \ref{fig:conf_mat_AffectNet7} shows the normalized confusion matrix of the $\text{AffectNet}_7$ on the test set. Multiplying these values with 100 will return the percentage of each predicted class by true class.

%\begin{figure}[!ht]
%    \centering
%    \includegraphics[scale=0.13]{figs/conf_mat_AffectNet7_v2.png}
%    \caption{$\text{AffectNet}_7$ Confusion Matrix  \label{fig:conf_mat_AffectNet7}}
%\end{figure}

\begin{figure}[!ht]
    \centering
    \includegraphics[scale=0.15]{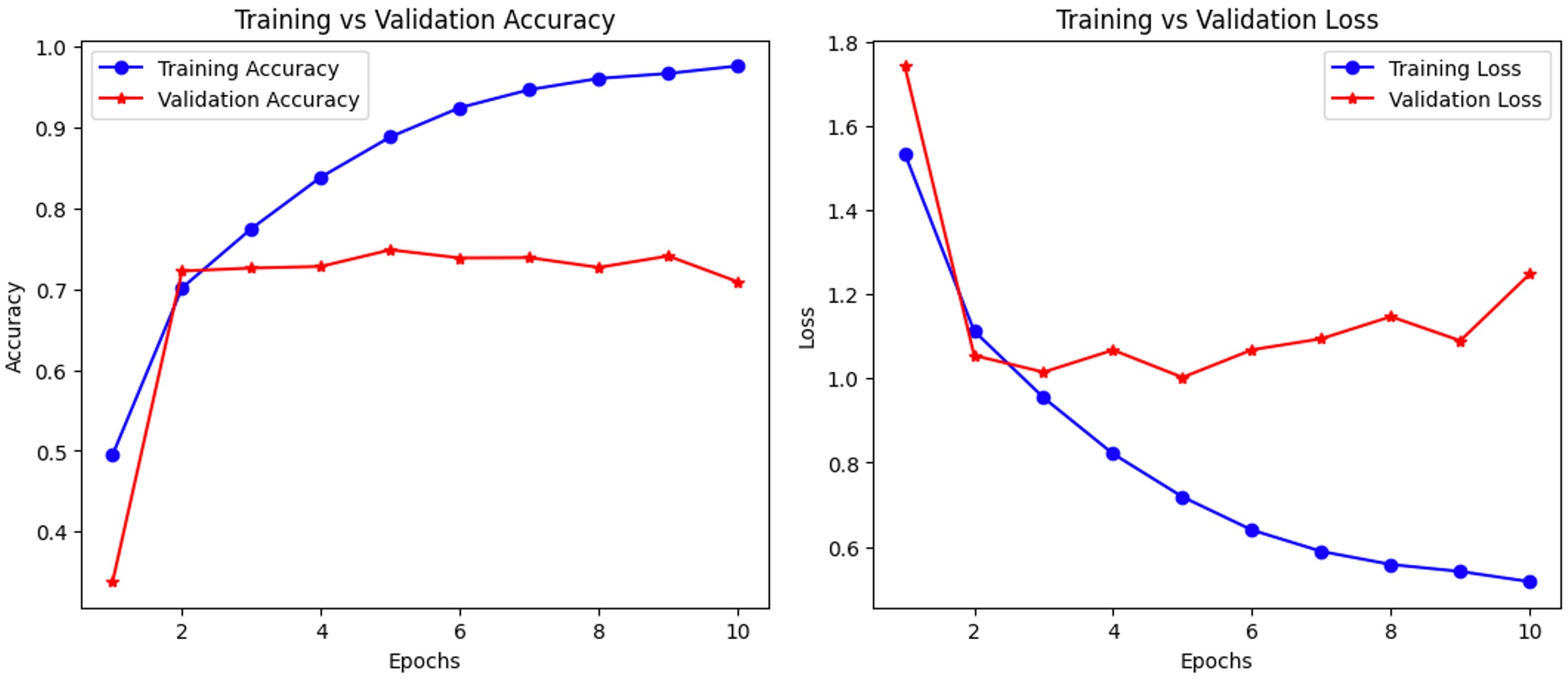}
    \caption{Training Accuracy/Loss vs Validation Accuracy/Loss: $\text{AffectNet}_7$.  \label{fig:epoch_loss_AffectNet7}}
\end{figure}

%In Table \ref{tab:test_acc_db}, evaluation results of our proposed ExpressNet-MoE model are listed in terms of accuracy and macro averaged precision, recall and $F_1$-score on the AffectNet (7-class) evaluation set.
The classification task on $\text{AffectNet}_7$ showed a robust learning progression. From 49.51\% in the first epoch to 97.60\% in the tenth (in Fig. \ref{fig:epoch_loss_AffectNet7} ). The training accuracy increased significantly. By the sixth epoch validation accuracy had reached 74.86\%, followed by slight variations that might indicate overfitting. Despite these differences, the final stored model matches an optimally generalized version that successfully captures emotional patterns in the $\text{AffectNet}_7$ dataset. The final test accuracy the model achieved in this dataset is 74.77\%. 

\subsection{$\text{AffectNet}_8$ Results}
%Fig. \ref{fig:conf_mat_AffectNet8} shows the normalized confusion matrix of the $\text{AffectNet}_8$ dataset for the test split.

\begin{figure}[!ht]
    \centering
    \includegraphics[scale=0.15]{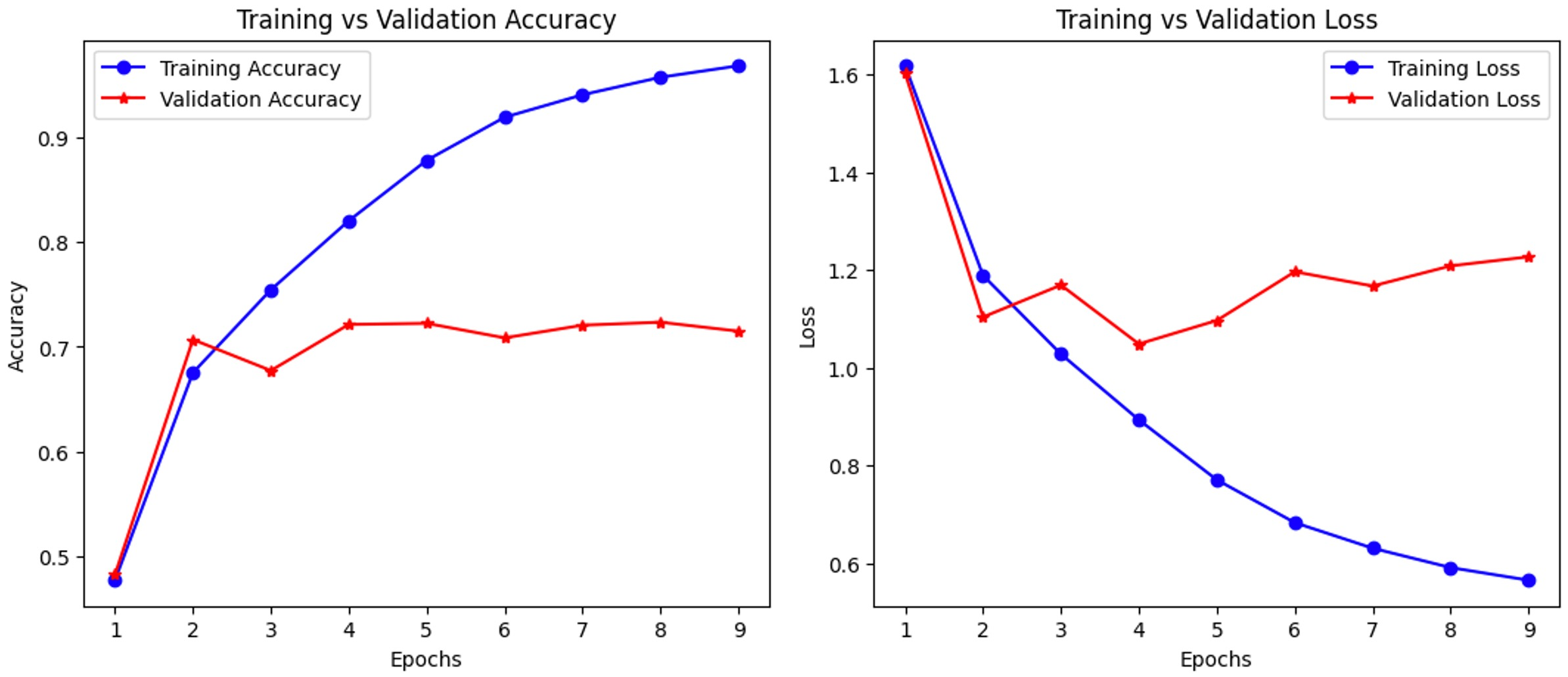}
    \caption{Training Accuracy/Loss vs Validation Accuracy/Loss: $\text{AffectNet}_8$.  \label{fig:epoch_loss_AffectNet8}}
\end{figure}

%\begin{figure}[!ht]
%    \centering
%    \includegraphics[scale=0.155]{figs/conf_mat_AffectNet8.png}
%    \caption{$\text{AffectNet}_8$ Confusion Matrix  \label{fig:conf_mat_AffectNet8}}
%\end{figure}
%Table \ref{tab:test_acc_db} also demonstrates the evaluation results of our proposed ExpressNet-MoE model in terms of accuracy and macro averaged precision, recall and $F_1$-score on the AffectNet (8-class) evaluation set.
Training accuracy increased significantly in the $\text{AffectNet}_8$ classification task, rising from 47.68\% in the first epoch to 96.88\% by the ninth (in Fig. \ref{fig:epoch_loss_AffectNet8}). 

While training accuracy kept increasing, suggesting possible overfitting, validation accuracy peaked at 72.35\% in the eighth epoch and then varied slightly. However, optimal generalization is ensured by the final stored model as it stores the model when highest validation accuracy is achieved. Even though the model was able to learn emotion representations, its resilience could be further increased by using additional regularization techniques. The final testing accuracy that the model achieved in this dataset is 72.55\%.

\subsection{RAF-DB Results}
%Fig. \ref{fig:conf_mat_RAF-DB}. shows the normalized confusion matrix of the RAF-DB dataset for the test split. 

%\begin{figure}[!ht]
%    \centering
%    \includegraphics[scale=0.13]{figs/conf_mat_RAF-DB.png}
%    \caption{RAF-DB Confusion Matrix  \label{fig:conf_mat_RAF-DB}}
%\end{figure}

%Table \ref{tab:test_acc_db} demonstrates the evaluation results of our proposed ExpressNet-MoE model in terms of accuracy and macro averaged precision, recall and $F_1$-score on the RAF-DB (7-class) evaluation set.

\begin{figure}[!ht]
    \centering
    \includegraphics[scale=0.15]{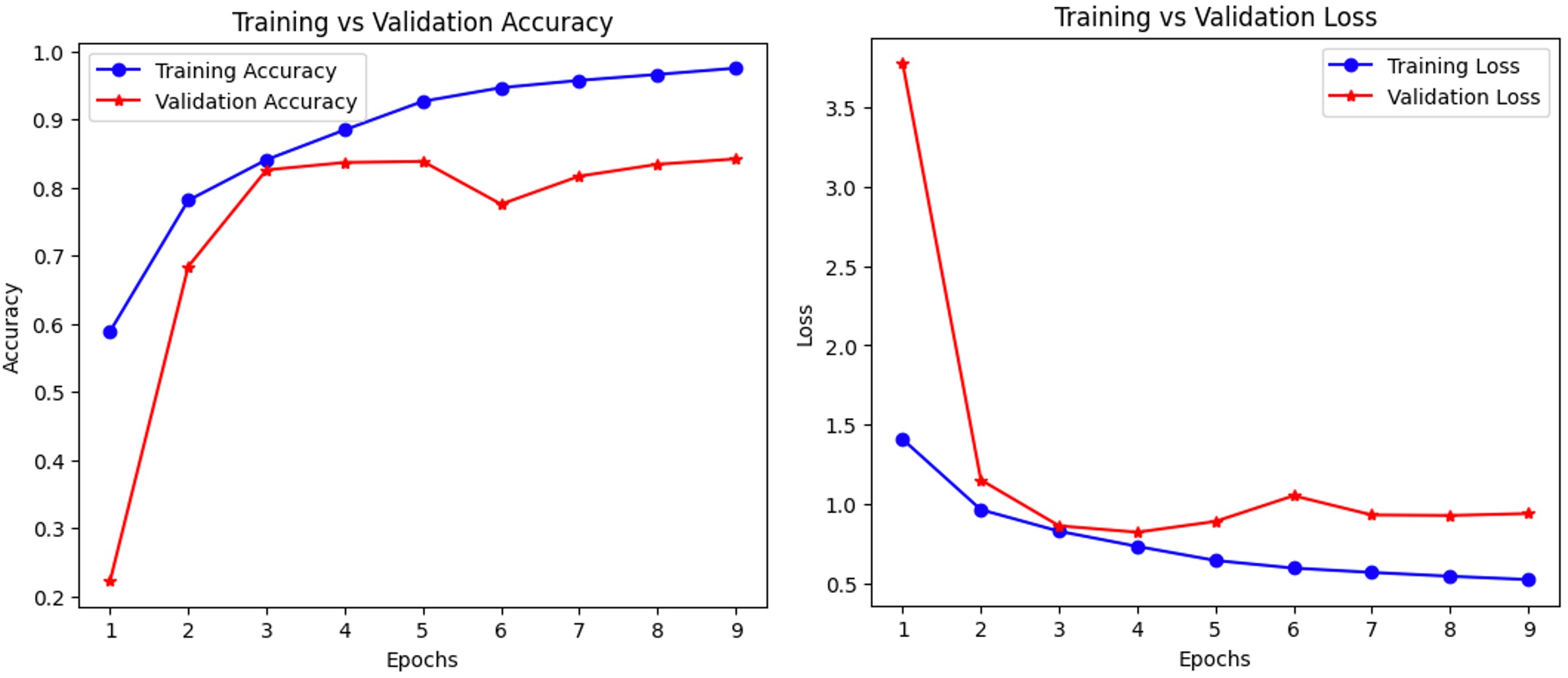}
    \caption{Training Accuracy/Loss vs Validation Accuracy/Loss: RAF-DB.  \label{fig:epoch_loss_RAF-DB}}
\end{figure}

Training accuracy increased from 58.89\% in the first epoch to 97.54\% by the ninth (in Fig. \ref{fig:epoch_loss_RAF-DB}), demonstrating a rapid improvement on the RAF-DB dataset. 

Similar trends were seen in validation accuracy, which first increased from 22.24\% to 84.21\% in the ninth epoch. The model eventually attained high generalization performance, despite a few slight decreases in the sixth and seventh epochs. Early stopping strengthened the model's stability for practical FER applications by preventing needless training past the point of peak performance. The final testing accuracy that the model achieved in this dataset is 84.29\%. 

\subsection{FER-2013 Results}
%Fig. 14. shows the normalized confusion matrix of the FER-2013 dataset on the test split.

%\begin{figure}[!ht]
%    \centering
%    \includegraphics[scale=0.62]{figs/FER-2013 Confusion Matrix.png}
%    \caption{FER-2013 Confusion Matrix  \label{fig:conf_mat_FER2013}}
%\end{figure}

\begin{figure}[!h]
    \centering
    \includegraphics[scale=0.15]{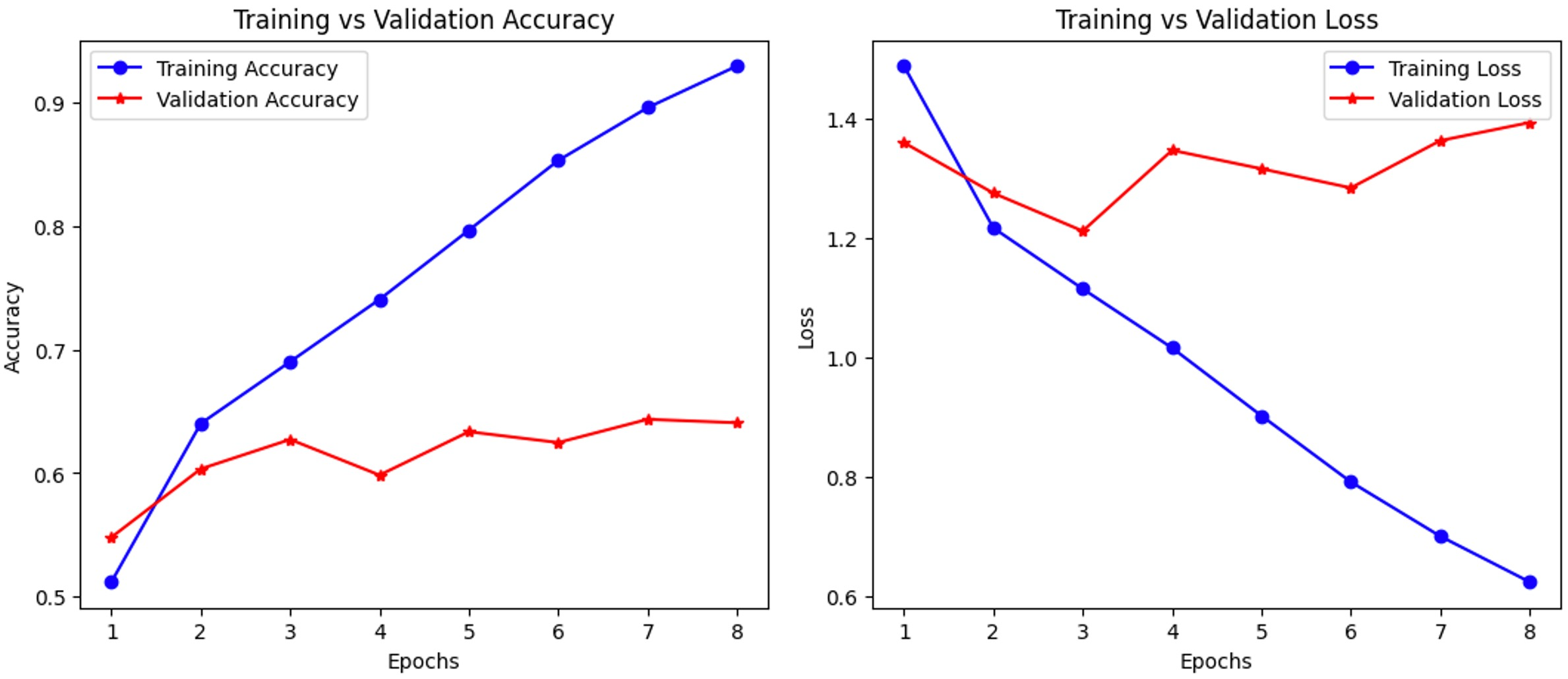}
    \caption{Training Accuracy/Loss vs Validation Accuracy/Loss: FER-2013.  \label{fig:epoch_loss_FER2013}}
\end{figure}

%Table \ref{tab:test_acc_db} demonstrates the evaluation results of our proposed ExpressNet-MoE model in terms of accuracy and macro averaged precision, recall and $F_1$-score on the FER-2013 (7-class) evaluation set.
A peak validation accuracy of 64.37\% and a training accuracy of 93.00\% are achieved on the FER-2013 training data (Fig. \ref{fig:epoch_loss_FER2013}). However, after peaking, validation accuracy fluctuated, mostly as a result of incorrectly categorized data in the dataset and overfitting. Despite this, the model was able to acquire strong emotion representations. Better preprocessing or label correction could improve the results even though the final stored model is the most generalizable version. The final testing accuracy that the model achieved in this dataset is 64.66%. 

In every dataset, the ``emotion'' happy is classified with the highest accuracy. This is due to the fact that all of the datasets had ``happy'' images predominantly.

\section{COMPARISON WITH STATE OF THE ART METHODS}
\label{sec:comparison}
Table \ref{tab:comparison} presents a detailed comparison of the ExpressNET-MoE model against other existing models on the $\text{AffectNet}_7$, $\text{AffectNet}_8$, RAF-DB, and FER-2013 datasets.

\begin{table}[tph]
    \centering
    \caption{Comparison against SOTA $\text{AffectNet}_7$: AN-7, $\text{AffectNet}_8$: AN-8, RAF-DB: RAF, FER-2013: FER) \label{tab:comparison}}
    \begin{tabular}{lcccc}
    \toprule
Model & AN-7	 & AN-8	& RAF & FER \\\midrule
ViT \cite{roka2023fine} & 64.48\% & - & - & -\\
MobileNet-v1 \cite{savchenko2022classifying} & 64.71\% & 60.25\% & - & -\\
EfficientNet-B0 \cite{savchenko2022classifying} & 65.74\% & 61.32\% & - & -\\
EmoNeXt-T \cite{bohi2024novel} & 65.55\% & 61.36\% & - & 73.34\%\\
EmoNeXt-S \cite{bohi2024novel} & 65.90\% & 62.51\% & - & 74.33\%\\
EmoNeXt-B \cite{bohi2024novel} & 66.21\% & 62.94\% & - & 74.91\%\\
EfficientNet-B2 \cite{savchenko2022classifying} & 66.34\% & 63.03\% & - & -\\
Face2Nodes \cite{jiang2023face2nodes} & 66.69\% & - & 91.02\% & -\\
EmoNeXt-L \cite{bohi2024novel}& 66.88\% & 63.12\% & - & 75.57\%\\
EmoNeXt-XL \cite{bohi2024novel} & 67.46\% & 64.13\% & - & 76.12\%\\
DDAMFN \cite{zhang2023dual} & 67.03\% & 64.25\% & 91.35\% & -\\
QCS \cite{wang2024qcs} & 67.94\% & 64.30\% & 92.50\% & -\\
DCNN-CBAM \cite{halim2023facial} & - & 66.09\% & - & 72.28\%\\
Norface \cite{liu2024norface} & - & 68.69\% & 92.97\% & -\\
D-CNN \cite{uniyal2024analyzing} & - & 70\% & - & -\\
D-CNN \cite{huang2023study} & 56.54\% & - & 83.37\% & -\\
GZS-ConvNet \cite{bhati2025generalized} & 59.61\% & - & 62.51\% & 75.32\%\\
FER-VT \cite{huang2021facial} & - & - & 88.26\% & -\\
ResEmoteNet \cite{roy2024resemotenet} & 72.93\% & - & 94.76\% & 79.79\%\\\midrule
ExpressNet-MoE (Ours) & 74.77\% & 72.55\% & 84.29\% & 64.66\%\\\bottomrule
    \end{tabular}
\end{table}

Our model outperforms all other models as mentioned in the table as it achieves an accuracy of 74.77\% on the $\text{AffectNet}_7$ dataset. The models that are its next competitors are ResEmoteNet (72.93\%), QCS (67.94\%) and EmoNeXt-XL (67.46\%), lag behind. Other models fall short too, including EfficientNet-B2 with an accuracy of 66.29\% and DDAMFN with an accuracy of 67.03\%. This substantial improvement in testing accuracy in our model demonstrates the effectiveness of our model in capturing nuanced facial emotions and fluctuations in $\text{AffectNet}_7$ which is a dataset renowned for its challenging real-world photographs. When compared to single-stream architectures such as EfficientNet and EmoNeXt, our methodology which incorporates multi-expert learning and employs ensemble CNN and ResNet-based feature extraction, has provided a superior representational capacity, which is evidenced by the substantial gap between our model and the preceding best-performing models.

Similarly, our model outperforms all prior approaches in classifying $\text{AffectNet}_8$ dataset by achieving 72.55\% as its testing accuracy on the dataset. For this dataset, D-CNN has the highest reported accuracy of 70\%, followed by Norface achieving 68.69\% and then QCS achieving 64.30\%. The margin of 2.55\% over D-CNN and 3.86\% over Norface underscores our model’s effectiveness in generalizing even when an additional emotion class (contempt) is included, which frequently complicates classification tasks due to its subtlety. Fine-tuned EfficientNet-B2 with an accuracy of 63.03\% and EmoNeXt-XL with an accuracy of 64.13\% perform significantly worse, emphasizing the advantages of our model's architecture for learning intricate facial emotions. The improvement in accuracy between $\text{AffectNet}_7$ and $\text{AffectNet}_8$ indicates that our model is both resilient to variations in the distribution of emotional classes and capable of managing large-scale datasets.

Our model achieves 84.29\% accuracy on the RAF-DB dataset which is competitive and proves the model’s adaptability and generalizability but still falls short of the top-performing techniques. ResEmoteNet achieves the highest accuracy of an accuracy of 94.76\%, followed closely by Norface with an accuracy of 92.97\% and QCS with 92.50\% accuracy. The models outperform our approach by approximately 8.68\% to 10.47\%. This can be attributed to the fact that RAF-DB is predominantly imbalanced and frequently requires fine-grained feature extraction. To enhance our ability to extract substantial representations from RAF-DB photographs, supplementary pretraining strategies along with data augmentation would be necessary.

Despite its performance on $\text{AffectNet}_7$, $\text{AffectNet}_8$ and RAF-DB, our model’s FER-2013’s testing accuracy of 64.66\% is inferior to that of the best-performing models. ResEmoteNet with an accuracy of 79.79\%, EmoNeXt-XL with 76.12\% accuracy, and GZS-ConvNet with an accuracy of 75.32\% all outperformed our method. FER-2013 presents significant challenges due to its grayscale nature and low-resolution photographs ($48\times 48$ pixels). Additionally, the dataset’s mislabeling introduces substantial errors. To enhance performance on FER-2013, state-of-the-art algorithms often employ additional preprocessing methods, transfer learning, or domain adaptation. While our method demonstrates effectiveness on AffectNet, it may not be as well-suited for grayscale photographs, potentially contributing to the observed performance decline. Furthermore, the high amount of noise and label ambiguity in FER-2013 may have an impact on models that depend on high-resolution features. It is worth noting that the majority of models achieving higher accuracies on FER-2013 exhibit poor performance on AffectNet due to presence of many mislabeled samples in it. %This could be attributed to the potential for training on mislabeled data and the memorization of wrong features.

\section{DISCUSSIONS}
\label{sec:discussion}
%The results of our study show that the ExpressNet-MoE model outperforms all existing state-of-the-art models. It achieves competitive accuracy across two central FER datasets, $\text{AffectNet}_7$ and $\text{AffectNet}_8$. 
Generalization and adaptability have been enhanced by our proposed model with the dynamic selection of the relevant experts, a capability enabled by the model’s mixture of experts (MoE) architecture. Furthermore, the transfer learning method and ensemble CNN architecture also adds to reducing the problems related to applying FER in real-world scenarios due to changes in light conditions, occlusion, and demographic diversity. %According to our findings, the model does not perform very well on the FER-2013 dataset. This could be attributed to the shortcomings of the FER-2013 dataset, including low resolution, mislabeled, and grayscale photographs. 
ExpressNet-MoE provides a robust means for FER applications in real-world scenarios, is capable of recognizing complex emotional expressions and provides a strong base for engagement detection systems as well by classifying emotional states more accurately.

Another important finding of this research is also that the model's capacity to identify some emotions, like happiness, is noticeably more accurate than other emotions like fear or disgust. This is attributed to the datasets' class imbalances, where certain emotions are overrepresented while others have few training samples. Even though the model benefits from adaptive feature learning, further improvements can be made involving including more complex data augmentation methods or fine-tuning methods using domain-specific datasets. This could increase the recognition of underrepresented emotions by FER models. Additionally, the performance difference between RAF-DB and high-performing models like ResEmoteNet indicates that fine-grained feature extraction may benefit from additional optimization. To improve the model's sensitivity to minute face changes, future research could investigate architectural changes such as adding attention processes.

\begin{figure}[!ht]
    \centering
    \includegraphics[scale=0.65]{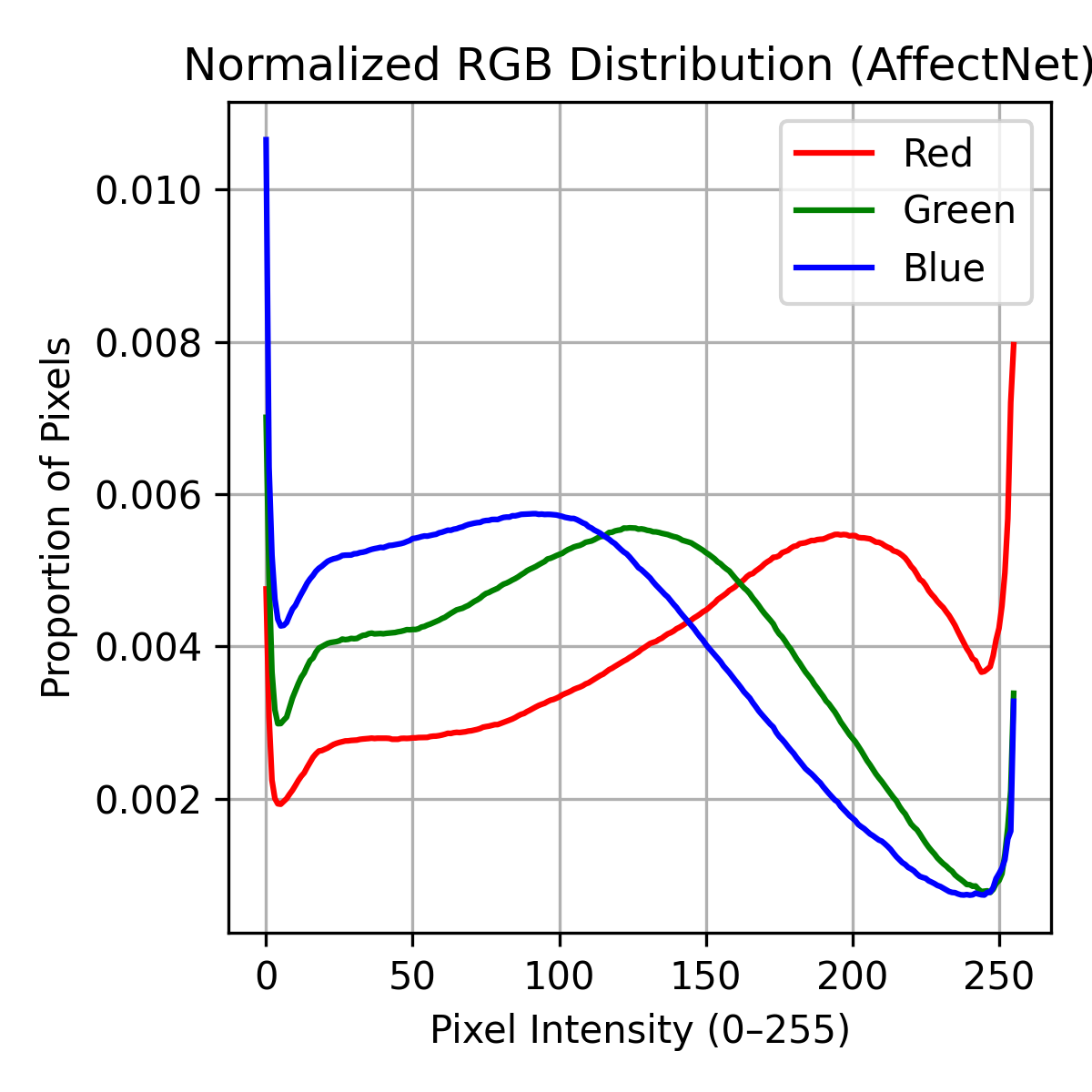}
    \includegraphics[scale=0.65]{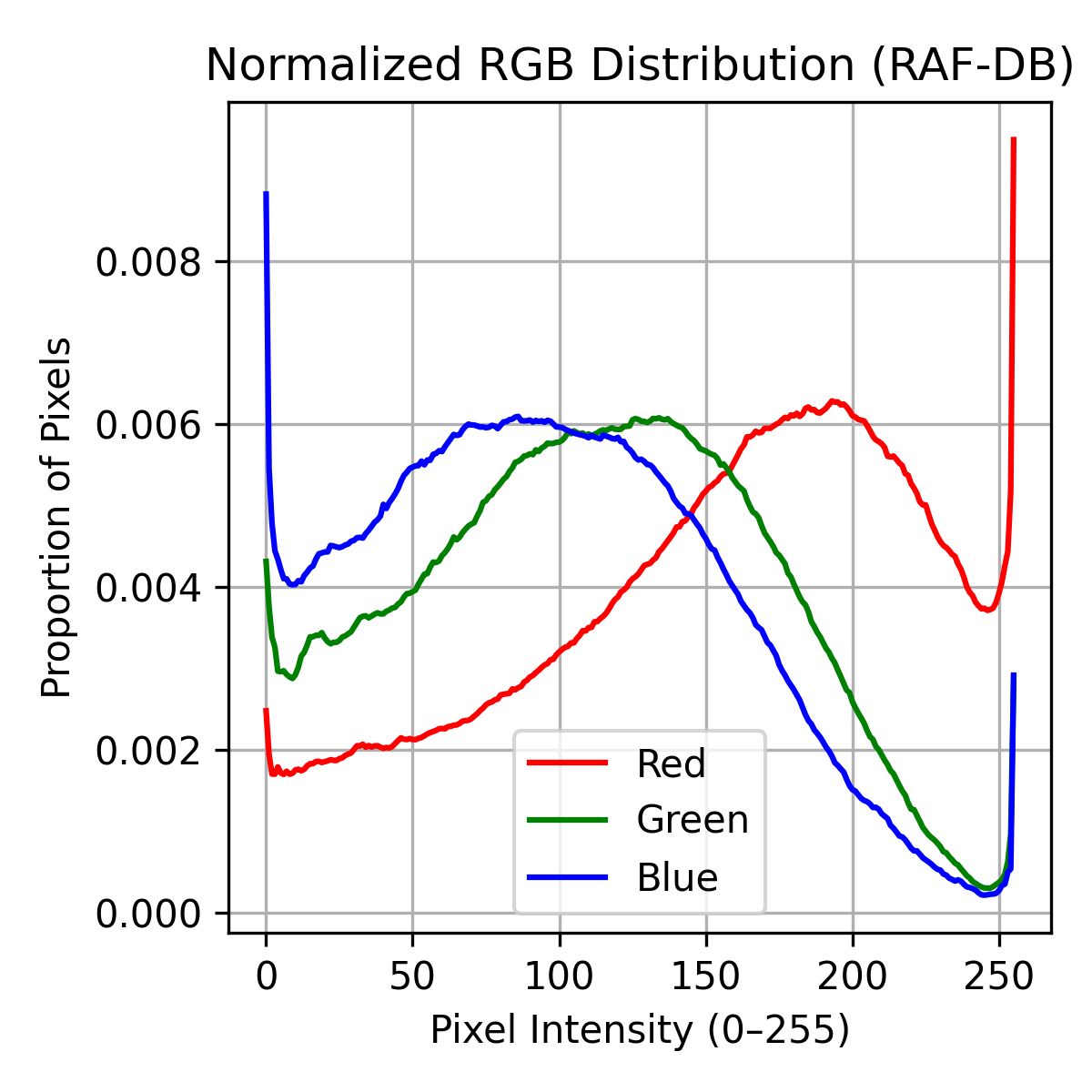}
    \caption{AffectNet (left) and RAF-DB (right) Normalized Color Distribution across all images\label{fig:ColorDist}}
\end{figure}

\begin{figure}[!ht]
    \centering
    \includegraphics[scale=0.75]{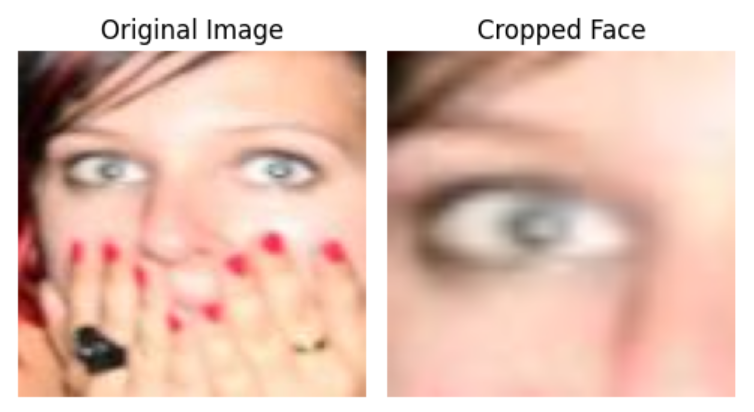}
    \caption{Mediapipe's Detection on RAF-DB\label{fig:RAF-DB-Cropped}}
\end{figure}

Furthermore, as we can see in Fig. \ref{fig:ColorDist}, the distribution of RGB values for AffectNet and RAF-DB are different with RAF-DB having higher pixel values for the red channel. This is attributed to variations in head position, occlusion, lighting conditions and skin color. The difference in the accuracies received for both RAF-DB and AffectNet can be attributed to this aspect of the datasets. We can also see in Fig. \ref{fig:RAF-DB-Cropped} that the Mediapipe library that we used for image pre-processing crops the entire face for many RAF-DB images as these images for emotions like surprise, fear, etc. have their faces covered. Therefore, the proposed architecture cannot properly classify these images for the RAF-DB dataset and underperforms for the dataset. 

\section{CONCLUSION}
\label{sec:conclusions}
%our model outperforms sota models for AffectNet and achieves competitive ??
To enhance facial emotion recognition, the proposed ExpressNet-MoE presents a novel hybrid deep learning architecture that successfully combines Mixture of Experts (MoE) module with Convolution Neural Networks and provides new insights into how expert models can be dynamically selected for emotion recognition tasks. This advances our knowledge of how ensemble approaches can be used to solve visual recognition problems, especially ones that involve intricate and subtle patterns like facial expressions. The model maintains competitive performance on RAF-DB while achieving excellent accuracy on $\text{AffectNet}_7$ and $\text{AffectNet}_8$ by utilizing dynamic expert selection and multi-scale feature extraction. The findings demonstrate the value of adaptive feature learning in addressing practical facial emotion recognition challenges including occlusion, lighting conditions, and demographic diversity of the subjects. ExpressNet-MoE lays a solid basis for future developments in such systems, despite certain limitations, especially in datasets with noisy labels or unbalanced class distributions. To improve the model's generalizability further in a range of applications, future research could concentrate on honing expert selection techniques or increasing the number of experts, improving image pre-processing as we see improper pre-processing degrades the model's performance, adding attention-based mechanisms, and increasing variance in training datasets.

%\section*{APPENDIX}
%Appendixes, if needed, appear before the acknowledgment.

\section*{ACKNOWLEDGMENT}
This material is based upon work supported by the National Science Foundation under Grant No. 2329919.

\bibliographystyle{plainnat}%{unsrtnat}
\bibliography{references}

\end{document}